\documentclass[12pt]{article}
\usepackage{arxiv}

\usepackage{graphicx}
\usepackage{subcaption}
\usepackage{amsmath}
\usepackage{xcolor}
\usepackage{amssymb}
\usepackage{lipsum}  
\usepackage{xspace}
\usepackage{comment}
\usepackage[inline]{enumitem}
\usepackage{stfloats}
\usepackage[pagebackref,breaklinks,colorlinks]{hyperref}

\newcommand{\tmodel}{\texttt{LEARN}\xspace}

\title{Latent Enhancing AutoEncoder for\\Occluded Image Classification 
}
\author{
Ketan~Kotwal\hspace{1mm}\mbox{\href{https://orcid.org/0000-0003-3766-0881}{\includegraphics[width=4mm]{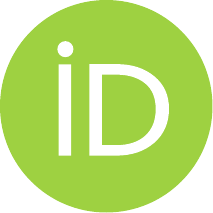}}}\\
\texttt{ketan.kotwal@idiap.ch}\\
\And
Tanay Deshmukh\\
\texttt{tanaydeshmukh96@gmail.com}\\
\And
Preeti~Gopal\hspace{1mm}\mbox{\href{https://orcid.org/0000-0002-5470-3911}{\includegraphics[width=4mm]{images/orcid.pdf}}}\\
\texttt{preetigopalindia@gmail.com}\\
}
\date{}

\makeatletter
\NewDocumentCommand\blfootnote{s O{12pt} m}{%
  \begingroup
  \renewcommand{\@makefntext}[1]{\noindent \IfBooleanT{#1}{\rule{0pt}{#2}}#3}
  \renewcommand\thefootnote{}\footnote{#1}%
  \addtocounter{footnote}{-1}%
  \endgroup
}
\makeatother

\hypersetup{
pdftitle={Latent Enhancing AutoEncoder},
pdfsubject={},
pdfauthor={Ketan Kotwal},
pdfkeywords={Biometrics, Occlusion Handling, Latent Enhancement.},
}

\begin{document}

\maketitle

\blfootnote*{\copyright{} 2024 IEEE. Personal use of this material is permitted. Permission from IEEE must be obtained for all other uses, in any current or future media, including reprinting/republishing this material for advertising or promotional purposes, creating new collective works, for resale or redistribution to servers or lists, or reuse of any copyrighted component of this work in other works.\\}

\begin{abstract}
Large occlusions result in a significant decline in image classification
accuracy. During inference, diverse types of unseen occlusions introduce
out-of-distribution data to the classification model, leading to accuracy
dropping as low as 50\%. As occlusions encompass spatially connected regions,
conventional methods involving feature reconstruction are inadequate for
enhancing classification performance. We introduce \tmodel: \textbf{L}atent \textbf{E}nhancing fe\textbf{A}ture \textbf{R}econstruction \textbf{N}etwork-- An auto-encoder
based network that can be incorporated into the classification model before its
classifier head without modifying the weights of classification model. In addition to reconstruction and classification losses,
training of \tmodel effectively combines intra- and inter-class losses
calculated over its latent space---which lead to improvement in recovering
latent space of occluded data, while preserving its class-specific
discriminative information. On the OccludedPASCAL3D+ dataset, the proposed \tmodel
outperforms standard classification models (VGG16 and ResNet-50) by a large
margin and up to 2\% over state-of-the-art methods. In cross-dataset
testing, our method improves the average classification accuracy by more than 5\% over the state-of-the-art methods. In every experiment,
our model consistently maintains excellent accuracy on in-distribution data\footnote{Supplementary Material can be downloaded from: \href{https://sigport.org/documents/supplementary-material-latent-enhancing-autoencoder-occluded-image-classification}{IEEE SigPort repository}}.

\end{abstract}


\section{Introduction}
\label{sec:intro}

Recent research has shown that despite their high performance in various areas,
deep convolutional neural networks (CNNs) encounter difficulties with occluded
data. This challenge presents significant obstacles in applications such as
autonomous driving~\cite{gilroy2019overcoming}, video
surveillance~\cite{kim2023people}, and medical imaging~\cite{zeng2021survey},
where precise classification of occluded objects is vital. Misidentifying
occluded objects can lead to accidents or decision-making errors, particularly
in the context of autonomous driving. Thus, addressing the issue of occlusions
is critical for strengthening the robustness and reliability of these models for
consistent and trustworthy deployment.

\begin{figure}[t!]
\centering
\includegraphics[width=0.6\columnwidth]{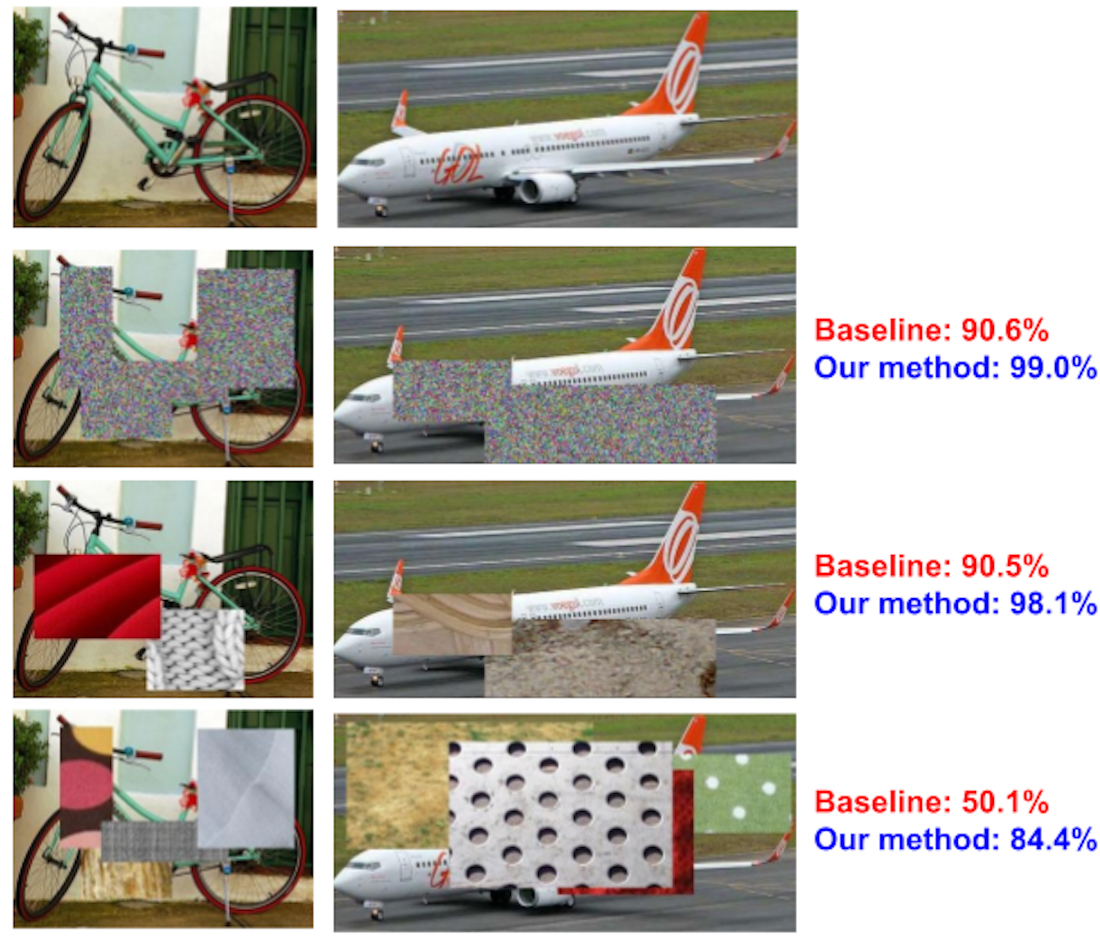}
\caption{Examples of clean and occluded images from OccludedPASCAL3D+ dataset. The clean images are at the top row, and their occluded versions with patches of noise (of level 5), texture (of level 5) and random objects (of level 9) are shown in subsequent rows respectively.
}
\label{fig:intro-image}
\end{figure}

The deterioration in performance of CNNs, (as shown in Fig.~\ref{fig:intro-image}), can be 
attributed to the complexity in handling occlusions that occur at different
scales, locations, and ratios~\cite{saleh2021occlusion}. Addressing all possible
occlusion patterns during training a classification model is extremely demanding
due to their vast number and variability~\cite{kortylewski2020combining}. 
For instance, the occluding object (\textit{occluder}) may be new and unseen in the training data.
Additionally, it can exist in any of the innumerable positions possible, and
occlude the target object (\textit{occludee}) to any extent (\textit{level}). Consequently,
deep learning-based models struggle to generalize well to
partially occluded objects.
The issue of occluded data can also be viewed from the perspective of
out-of-distribution (OOD) data. Occlusions bring about significant visual
variation, leading to shifts and deviations in the underlying data distribution
compared to the training dataset. Therefore, CNN models trained on specific
datasets often face challenges in accurately classifying occluded objects
because they have not been exposed to such variations during training.

The improvement towards resilience to OOD data is a significant
area of study within deep learning and computer vision. This focus has been
recognized through specialized challenges and competitions, such as the OOD-CV
challenge series at ICCV~2023\footnote{https://www.ood-cv.org} and
ECCV~2022\footnote{https://www.ood-cv.org/2022/index.html}. Recent benchmark
studies~\cite{zhao2022ood, zhao2023ood} categorize the OOD problem based on
nuisance factors such as illumination, pose, context, weather variations, and
occlusions. These studies indicate that among nuisance factors, occlusion presents
the most demanding challenge where classification accuracy for a ResNet-50
architecture~\cite{he2016deep} model drops by up to 25\% on the OOD-CV-v2
benchmark dataset. Consequently addressing occluded data remains pivotal in
enhancing robustness and reliability of deep CNN-based classification models.

In this work, we tackle the problem of classifying objects that are subject to
various levels and forms of occlusions. The key to accurately classifying an
occluded object lies in properly reconstructing the underlying latent space that
ascribes the correct class. Moreover, it is crucial that any approach for
handling the OOD data has minimal impact on images of non-occluded objects
(\textit{i.e.}, should not degrade in-distribution data). This challenge is akin
to removing noise from large and connected components of complex class-specific
features obtained from deep CNNs. We harness the inherent capability of
autoencoders (AE) to acquire robust and compact representations of data.
We design an AE-based deep network-- \tmodel (\textbf{L}atent \textbf{E}nhancing  fe\textbf{A}ture \textbf{R}econstruction \textbf{N}etwork)-- that not only learns to enhance
specific classes' latent features but also reconstructs occluded features. This
is accomplished by training \tmodel on samples containing both occluded and
non-occluded (clean) objects from corresponding classes. For this purpose, we create occluded images with
various types and levels of occlusion on-the-fly by randomly occluding different
parts of the clean images, mimicking real-world scenarios. As occlusions may
hinder as much as 70+\% of the spatial area (see last two rows of Fig.~\ref{fig:intro-image}), it is extremely difficult to recover
the true (or clean) features solely by observing the local neighborhood. Hence,
we incorporate auxiliary loss terms which constrain the similarity between the
latent space of occluded and clean images while maintaining sufficient
inter-class distances within latent space. Lastly, we utilize classification
loss (through frozen fully connected (FC) layers) to supervise \tmodel's training and align it
with the backbone model. We do not modify any parameters of the backbone model
(apart from the final FC layer finetuned to given number of classes), allowing for easy integration of our proposed
\tmodel into inference pipelines with minimal adjustments necessary.

The contributions of our work can be summarized as
follows:
\begin{itemize}[noitemsep,topsep=0pt,parsep=0pt,partopsep=0pt, leftmargin=*]
\item We design an auto-encoder based network that can be incorporated within a classification CNN to enhance its robustness at classifying
objects with varying degrees and types of occlusions. It can seamlessly work across different CNN classification models.

\item We propose auxiliary loss functions that facilitate efficient reconstruction of occluded data in the latent space while simultaneously enhancing their discriminative capability.

\item Our results demonstrate that the \tmodel model leads to significant
improvements in different types of occluded images, and outperforms recent
state-of-the-art methods. It also maintains excellent accuracy for clean,
non-occluded images. With only 0.7M parameters, the \tmodel for VGG16 backbone improves the classification accuracy from 55 \% (baseline) to 86\% for occlusions as high as 60--80\% of object area. For the same set of occlusions, the accuracy of classification is improved by 28\% over the baseline and 3\% over state-of-the-art methods for a ResNet-50 backbone with 2.5M parameters. 
\end{itemize}

%
\section{Related Work}
\label{sec:related_work}

Although the deep CNN-based classification models are highly discriminative~\cite{alexnet,simonyan2015deep}, their performance rapidly deteriorates with increasing level and complexity of occlusions. On the other hand, compositional models are more robust~\cite{deepvoting} since they represent objects as a composition of various spatial components of the image. Hence their overall inference is accurate even when a few spatial components are wildly different due to partial occlusions. In~\cite{kortylewski2020combining}, the classical CNN model and a compositional mixture model are stacked together leveraging the advantages of both networks. Here the compositional model is trained with the feature vectors in the final convolutional layer, making the model independent of attributes such as illumination and pose. During inference, the image is first passed through the CNN. Only when this module classifies with low certainty indicating the possible presence of occlusions, the features are fed to the compositional model which predicts the final class. In an extension of this work, Kortylewski \textit{et al}.\ introduced Compositional Neural nets wherein the final fully connected classification layer is replaced by a differentiable composition of mixture models~\cite{Adam2021}. The parameters of the mixture model were trained such that for each category of object, one of the mixture models gets activated. The Compositional Neural Net was further modified in \cite{Wang2020} to learn separate representations of the object and its background (context). In cases of strong occlusions, this prevents the model from falsely detecting the object based on the features of the context. This strategy combined with an estimation of the object's bounding box led to improved performance on severely occluded vehicle images from MS-COCO~\cite{mscoco} and PASCAL3D+~\cite{pascal3dplus} datasets. 

Another way of handling occlusions is by introducing better augmentation strategies in order to introduce better generalization to the models. In this regard, the idea of Soft Augmentation was recently introduced in \cite{SoftAug2023}, wherein the confidence in target label was reduced non-linearly based on the level of augmentation of the training sample. This led to better object classification under occlusions. However, the reported occlusion types were synthetic consisting of box-shaped constant intensity patches, mimicking the original image being cropped to different degrees. The performance of this strategy on images with realistic occluders is yet to be explored. 


\noindent\textbf{Available Datasets}: The availability of standard annotated occluded images is limited. Simple occluders usually consist of a constant gray-scale mask, patches of white noise or textures, while realistic occluders are real-life objects themselves. There are limited number of datasets of the latter kind. As an example, in OccludedPASCAL3D+ dataset (hereafter referred to as Pascal)~\cite{Wang2020}, occlusions were generated by superimposing objects from MS-COCO~\cite{mscoco} onto objects in PASCAL3D+~\cite{pascal3dplus} images. Zhan \textit{et al}.\ established an automatically generated subset of images from the COCO dataset with partial occlusions~\cite{occludedcoco}. In multiple other works, task-specific occlusions are created in specific ways for demonstrating the efficacy of the corresponding methods. For example, in~\cite{Feng2021}, a small reasonable number of clean and occluded image pairs are assumed to be available. Given a pair, a Deep-Feature-Vector (DFV) is extracted from each image. 
The difference between the DFV of clean and occluded image has information about the occluder. This difference vector is then added to the DFVs of all other clean images which do not have an occluded pair for the purposes of training. Training with such perturbed DFVs improved classification performance under occlusions. Other datasets for specialized tasks include the Real World Occluded Faces (ROF) 
dataset~\cite{Erakin2021} created specifically for face recognition.

%
\section{\textsc{\tmodel} for Occlusion-Robust Classification}
\label{sec:proposed_method}

We first provide overall functioning of the proposed \tmodel model, and also discuss the distinction between the current task and basic denoising or reconstruction problems. Following that, we explain the design choices and working of our loss functions towards training of the \tmodel.

Let $f \in \mathbb{R}^{c.h.w}$ be the output of the final convolutional layer of
the classification CNN, which we also refer to as the backbone network. In a
typical CNN, the corresponding output, $f$, is flattened and passed
through one or more FC layers, where the final FC layer has
dimensionality equal to the number of classes and its output
indicates the likelihood of the input belonging to each class. In the following
discussion, we denote $x_{m, n}$ as the $m$-th sample of the $n$-th class, and
its corresponding features as $f_{m, n}$. Additionally, the occluded versions of
these samples and corresponding features are denoted as $x_{m, n}^\text{occ}$
and  $f_{m, n}^\text{occ}$, respectively.

\begin{figure*}[!t]
\centering
\begin{subfigure}[b]{0.65\textwidth}
    \includegraphics[width=\textwidth]{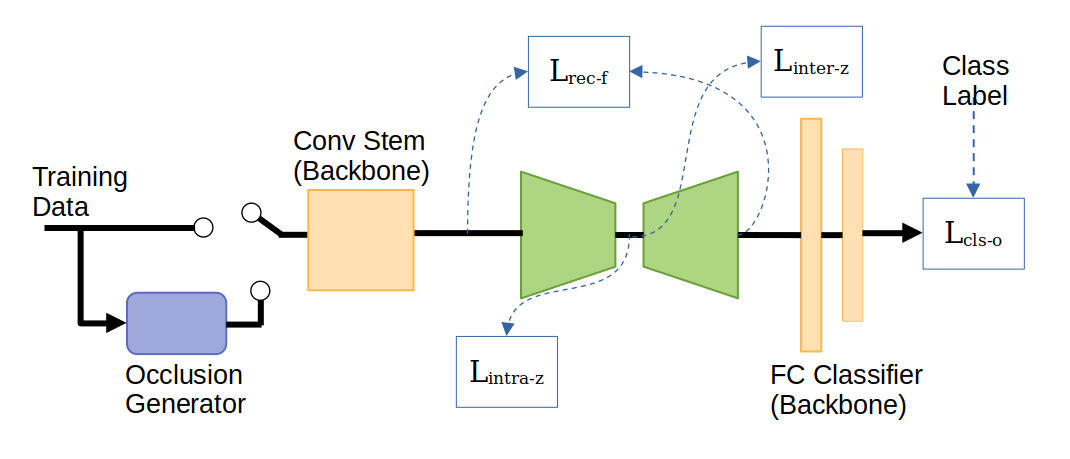}
    \caption{}
\end{subfigure}
\quad
\begin{subfigure}[b]{0.3\textwidth}
    \includegraphics[width=\textwidth]{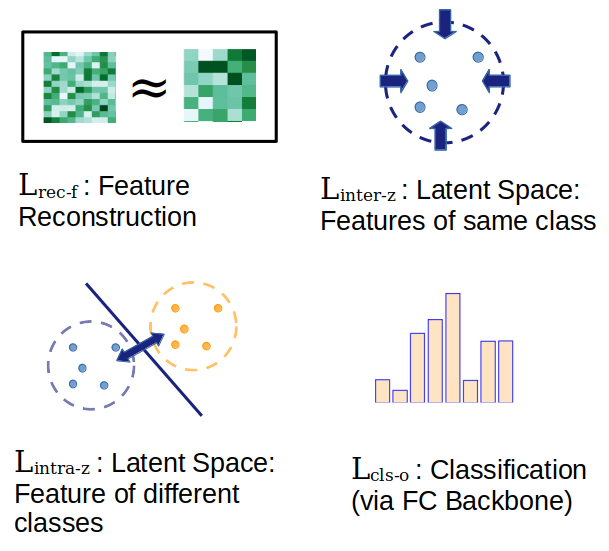}
    \caption{}
\end{subfigure}
\caption{The schematic of the proposed \tmodel: (a) shows the overall training pipeline along with loss functions, and (b) provides simple illustration of individual loss components. The green block depicts \tmodel in the form of AutoEncoder.}
\label{fig:ood_method}
\end{figure*}

To improve robustness to occlusions, working with feature representation $f$ is
often more effective than than directly working with images
\cite{kortylewski2020combining}. Likewise, we regard these feature maps,
$f$, as the input for our AE-based \tmodel where the goal is to learn the latent
space for a set of inputs of one or more classes and also to reconstruct the
features of occluded images for improved classification. As shown in Fig.~\ref{fig:ood_method}, the reconstructed
features are then fed back to the cascade of FC layers of the backbone network for
further processing. Therefore, one may view our proposed \tmodel being integrated
into an existing pretrained classification backbone for the purpose of correcting
or enhancing the features originating from occluded data. It is important to emphasize that
enhancing features of occluded data may seem like a
denoising problem, but significantly differs from processing images corrupted
with noise such as AWGN or salt-and-pepper. These forms of noise affect the
local neighborhood of pixels, whereas occlusion substantially degrades large 
spatial regions-- thereby making it nearly impossible to reconstruct or learn from
local-level interactions of pixels/ features. Consequently, a conventional denoising AE, trained with simple
reconstruction loss, does not yield a satisfactory solution to OOD
classification tasks. 

For a given \tmodel, we hypothesize that a joint approach to reconstruction
and classification is crucial for successful classification of occluded data.
It is necessary to reconstruct the features of occluded data in such a way that they
can serve as improved inputs for classification-- without compromising the accuracy
on clean, non-occluded data. We propose a following multi-objective loss
function to train the \tmodel for our classification task.

\noindent\textbf{Architecture of the \tmodel:} We use a conventional AE architecture where the encoder consists of a stack of 3 convolutional layers (conv) with the kernel size of 3. The number of output channels for each conv  layer are 64, 64, and 32, respectively. Except the last conv layer (which leads to the latent space), outputs of other conv layers are passed through ReLU activation and pooling. The latent vector is passed through the decoder of the \tmodel to reconstruct the features. The decoder architecture is exactly mirrored from that of the encoder by replacing conv by transposed-conv operations. The outputs of final transposed-conv layer are restricted to [-1, +1] through hard Tanh activation.

\noindent\textbf{Reconstructing Occluded Features:} The main objective of the \tmodel is an
attempt to reconstruct the features of occluded image. Initially, we create occlusions on a clean image,
as a part of preprocessing and input the corresponding features ($f_{m,
n}^\text{occ}$) for training the \tmodel. The features of clean image ($f_{m, n}$) act
as the reference or ground truth for reconstruction using MSE loss as shown by
Equation~\ref{eq:loss_rec_occ}.

\begin{equation}
\mathcal{L}_{\text{rec-f}}  = \bigl|\bigl| \tilde{f_{m, n}^\text{occ}} - f_{m, n} \bigr|\bigr|_{2}^{2},
\label{eq:loss_rec_occ}
\end{equation}
where $\tilde{f}$ indicates the reconstructed features, \textit{i.e.}, the output of \tmodel.

\noindent\textbf{Constraining Intra-Class Latent Space:} Due to occlusion's substantial 
impact on large regions 
of image/pixel data, majority of underlying signal (pixel data) is missing. The
current trend for addressing such challenges involves the use of generative networks; however, these
require massive amounts of training data and computational resources. 
We simplify this challenge by focusing on not attaining perfect reconstruction but instead ensuring that the reconstructed features contain sufficient class-discriminatory information.
As a result, we define an auxiliary loss function to ensure that
the latent space of \tmodel does not alter while processing different samples of
the same class, but enforces the latent vector of occluded samples to lie close to that of clean samples. For samples $x_{m1, n}$ and $x_{m2, n}$,  we achieve these objectives by calculating the loss on pairs of clean-clean as well as clean-occluded images. The corresponding loss $\mathcal{L}_{\text{intra-z}}$ is the MSE between the corresponding latent vectors $z$, provided as:
\begin{equation}
\mathcal{L}_{\text{intra-z}}  = \bigl|\bigl| z_{m1, n} - z_{m2, n} \bigr|\bigr|_{2}^{2}.
\label{eq:loss_2}
\end{equation}

\noindent\textbf{Discriminating Inter-Class Latent Space:} 
Given the significant diversity in occluded data, the \tmodel is vulnerable to learn
non-compact, over-complete latent space for individual classes---which may
degrade the classification accuracy. To address this issue, we introduce auxiliary 
inter-class latent loss that enforces discriminability among
samples of different classes within latent space. This loss,  $\mathcal{L}_{\text{inter-z}}$, formulated in a contrastive framework, is calculated on both- features of clean and occluded images of
different classes as shown in Equation~\ref{eq:loss3}.
\begin{equation}
\label{eq:loss3}
\mathcal{L}_{\text{inter-z}} = y\, (z_{m, n1} -z_{m, n2})^2 + (1-y) \, \max(0, M - (z_{m, n1} -z_{m, n2})^2), 
\end{equation}
where $y$ indicates a class-similarity label, which is set to 1 for same
classes, \textit{i.e.} $n1 == n2$, else zero; while \textit{M} indicates a
permissible inter-class distance in the latent space.

\noindent\textbf{Classification Loss:} Finally, we employ the FC-layer cascade from the
backbone model, along with its learnt weights, to improve the performance of the
\tmodel towards classification. The output of \tmodel is passed to the
frozen FC layer(s) to obtain a $D$-dimensional vector which is then used
to compute the cross-entropy classification loss. The loss is backpropagated to
the \tmodel without altering the FC layers of backbone. In addition to improving
the classifiability of occluded data, this loss term guides alignment of
reconstructed features with the pretrained classifier of the backbone. If
$\Theta(.)$ denotes the parameters of the FC-layers of the backbone, $d \in D$
refers to the class label, and \textsl{CE} is the standard cross-entropy loss,
the classification loss $\mathcal{L}_{\text{cls}}$ term can be obtained as:

\begin{equation}
\mathcal{L}_{\text{cls-o}}  = \text{CE}\left(\Theta(\tilde{f_{m, n}^\text{occ}}, d)\right),
\label{eq:loss_4}
\end{equation}
We compute $\mathcal{L}_{\text{cls-o}}$ on features of both- occluded and clean data.

The overall loss function for training the \tmodel is obtained by a weighted combination of
the aforementioned loss components as shown in Equation~\ref{eq:loss_total}. The
variables $\lambda_{-}$ refer to the relative weight of corresponding loss term.
\begin{equation}
\mathcal{L}_{\text{total}} = \mathcal{L}_{\text{rec-f}} + \lambda_{\text{intra-z}} \mathcal{L}_{\text{intra-z}} + \lambda_{\text{inter-z}} \mathcal{L}_{\text{inter-z}} + \lambda_{\text{cls-o}} \mathcal{L}_{\text{cls-o}}
\label{eq:loss_total}
\end{equation}
The combined loss term enables multiple strategies to synergistically address occluded features and limit the impact on latent space without compromising performance on high-quality (clean) data.

%
\section{Experimental Results}
\label{sec:expt_results}
\noindent\textbf{Datasets:} We evaluate the efficacy of \tmodel on two datasets
well-referenced for this task: the Pascal and MS-COCO Occluded Vehicles dataset
(hereafter referred to as MS-COCO). Pascal dataset is a subset of the PASCAL3D+
dataset \cite{pascal3dplus}, created by extracting vehicle images and
occluding them synthetically by four types of occluders: white noise, random
noise, textures and lastly objects. Depending on the percentages of occluded
area, the levels of occlusion can be classified into four categories: L0 (0\%),
L1 (20--40\%), L2 (40--60\%), L3 (60--80\%). Whereas occlusions in the Pascal
dataset are synthetic, we also consider evaluation of the proposed \tmodel on
the MS-COCO dataset~\cite{kortylewski2020compositional} that contains real
occlusions. Throughout this work, we have used the training split of Pascal
dataset to train our models.

\noindent\textbf{Backbones:} For our experiments, we consider two backbone
architectures, ResNet-50 and VGG16-- which are pre-trained on the ImageNet-1k
dataset. To obtain baseline results, we first fine-tune the backbones on the
training partition of the Pascal dataset (thus, 12 classes here), following the setup described in \cite{wang2015unsupervised}. 

\noindent\textbf{\tmodel Pipeline:~\footnote{The code to reproduce experiments from this paper is available at (the link will be shared after acceptance of the paper)}} The finetuned backbones (baseline models)
act as the reference for our subsequent experiments. We incorporate the
proposed AE-based \tmodel after the final convolutional layer, as shown in
Fig.~\ref{fig:ood_method} while freezing the weights of all backbone layers. To
train the \tmodel to reconstruct the feature maps of the occluded images, we
generate comprehensive occlusions on-the-fly where we occlude the training
samples using four occluders: white noise, random noise, texture, natural
objects and the degree or level of occlusions varies from 10\% to 90\%. For
each backbone: VGG16 and ResNet-50, we trained the model for 40 epochs  (with
early stopping criteria) with an initial learning rate of $1e^{-4}$ and batch
size of 128.

\begin{table}[t]
\centering
\renewcommand{\arraystretch}{1.3}
\resizebox{0.95\linewidth}{!}{
\begin{tabular}{|l|c|cccc|cccc|cccc|c|}
\hline
Occ. Area & \textbf{L0: 0\%} & \multicolumn{4}{c|}{\textbf{L1: 20--40\%}} & \multicolumn{4}{c|}{\textbf{L2: 40--60\%}} & \multicolumn{4}{c|}{\textbf{L3: 60--80\%}} & \textbf{Mean}
\\ \hline
Occ. Type & - & w & n & t & o & w & n & t & o & w & n & t & o &                     \\ \hline
{baseline} & 99.9 & 98.2 & 97.6 & 97.9 & 94.7 & 94.1 & 90.6 & 90.5 & 72.2 & 69.8 & 53.2 & 50.1 & 48.1 & 81.3 \\
CoD* & 92.1 & 92.7 & 92.3 & 91.7 & 92.3 & 87.4 & 89.5 & 88.7 & 90.6 & 70.2 & 80.3 & 76.9 & 87.1 & 87.1 \\
VGG+CoD* & 98.3 & 96.8 & 95.9 & 96.2 & 94.4 & 91.2 & 91.8 & 91.3 & 91.4 & 71.6 & 80.7 & 77.3 & 87.2 & 89.5 \\
TDAPNet* & 99.3 & 98.4 & 98.6 & 98.5 & 97.4 & 96.1 & 97.5 & 96.6 & 91.6 & 82.1 & 88.1 & 82.7 & 79.8 & 92.8 \\
CompNet* & 99.3 & 98.6 & 98.6 & 98.8 & 97.9 & \textbf{98.4} & 98.4 & 97.8 & 94.6 & \textbf{91.7} & 90.7 & \textbf{86.7} & 88.4 & \textbf{95.4} \\ \hline
Proposed & \textbf{100} & \textbf{99.7} & \textbf{99.8} & \textbf{99.6} & \textbf{99.0} & 98.3 & \textbf{99.0} & \textbf{98.1} & \textbf{96.1} & 80.5 & \textbf{91.9} & 84.4 & \textbf{89.3} & 95.1 \\ \hline
\end{tabular}
}%
\caption{Performance evaluation of \tmodel on the Pascal dataset, using VGG16 backbone, on varying levels and types of occlusions. Occlusion level signifies the percentage of the input image that is occluded and the occlusion types are, w=white noise, n=random noise, t=texture and o=natural objects. All values refer to classification accuracy in \%. For the methods marked with *, we report the results from \cite{kortylewski2020compositional}.}
\label{table:results_pascal_vgg}
\end{table}

\noindent\textbf{Results on Pascal:} Table \ref{table:results_pascal_vgg} shows
the performance, in terms of classification accuracy, of proposed \tmodel along with comparison with some of the recent state-of-the-art methods such as
CompositionalNets~\cite{kortylewski2020compositional},
dictionary-based-compositional model~\cite{kortylewski2020combining},
TDAPNet~\cite{TDAPNet}. We omit the details of other methods due to
brevity of space. We also include the classification accuracy on the baseline
(\textit{i.e.} the backbone finetuned on the clean Pascal dataset). For all
experiments, we maintain consistency across train and test protocols used by
other comparative methods---which facilitates straightforward comparison.

It can be observed that the proposed \tmodel outperforms CoD, VGG+CoD and 
TDAPNet methods under every occlusion type and level with improvements as high
as 10\% for severely occluded images. Compared to the baseline, \tmodel
achieves an improvement of 13.8\% in terms of average classification accuracy
(consists of clean + OOD data). Using \tmodel brings about 1--2\% improvement
across most occlusion types/levels compared to state-of-the-art CompNets,
though, on average, the proposed method falls short of CompNets by 0.3\%. One 
of the most challenging scenario is presented by occluding the image by another
object (\textit{e.g.} lower rows of Fig.~\ref{fig:intro-image}). For these
occlusions, \tmodel outperforms every competitive method across different
levels of occlusions (referred under the columns \textbf{o} in
Table~\ref{table:results_pascal_vgg}). It is important to note that the
incorporation of proposed model into VGG16 backbone does not degrade the
performance on clean images (level L0). While other methods resulted in nominal
decrease in classification accuracy of clean images, \tmodel yielded 
perfect classification of the same.

The results of our experiments with ResNet-50 backbone are provided in 
Table~\ref{table:results_pascal_resnet}. We also enlist the results of recent
state-of-the art methods which include RCNet \cite{jesslen2023robust} with a
3D-aware head, and CompNet~\cite{kortylewski2021compositional}. In terms of
average classification accuracy, \tmodel improves the baseline by 14.2\%.
On Pascal test dataset, the \tmodel outperforms all the state-of-the art
methods implementing a ResNet-50 backbone by margins ranging from 1.8--8.3\%.
The consistent improvement in classification across all levels of occlusion
demonstrates efficacy of the proposed \tmodel across different architectures of
classification CNNs.

\begin{table}[t]
\renewcommand{\arraystretch}{1.3}
\centering
\begin{tabular}{|l|cccc|c|}
\hline
Occ. Area & \textbf{L0} & \textbf{L1} & \textbf{L2} & \textbf{L3} & \textbf{Mean} \\ \hline
baseline & 99.9 & 94.5 & 83.0 & 55.4 & 79.4 \\
RCNet \cite{jesslen2023robust} & 99.1 & 96.1 & 86.8 & 59.1 & 85.3 \\
RCNet++ \cite{jesslen2023robust} & 99.4 & 96.8 & 87.2 & 59.2 & 85.7 \\ 
CompNet-Res50 \cite{kortylewski2021compositional} & 99.3 & 98.2 & 94.9 & 80.5 & 91.8 \\ \hline
Proposed & \textbf{100} & \textbf{99.3} & \textbf{96.3} & \textbf{83.5} & \textbf{93.6} \\ \hline
\end{tabular}
\caption{Performance evaluation of the \tmodel on the Pascal dataset, using ResNet-50 backbone, on varying levels of occlusions. All values refer to classification accuracy in \%.}
\label{table:results_pascal_resnet}
\end{table}

\begin{figure}[h]
    \centering
    \begin{subfigure}{0.35\columnwidth}
        \centering
        \includegraphics[width=\textwidth]{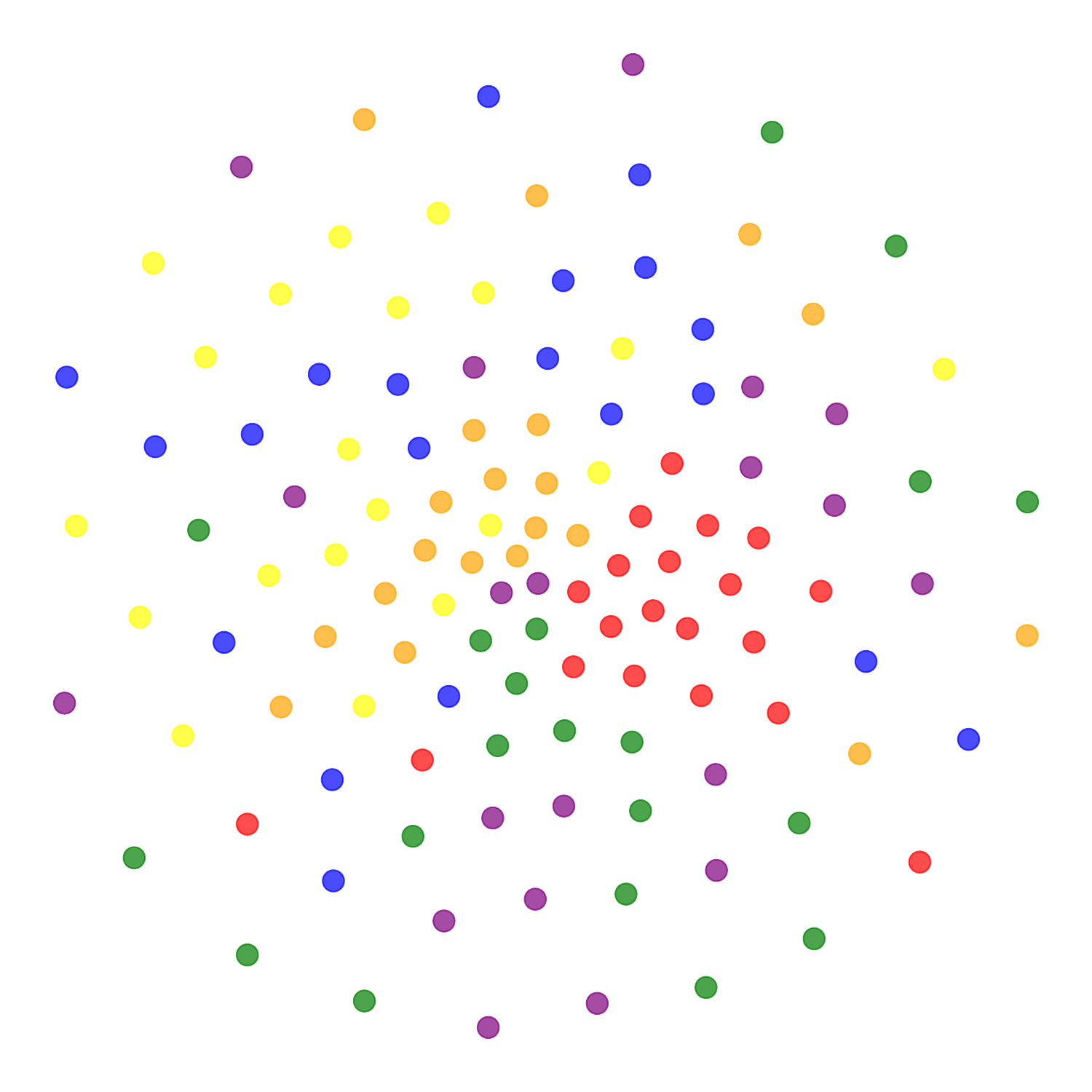}
        \caption{Baseline}
        \label{fig:tsne_baseline_vgg}
    \end{subfigure}
\qquad \qquad
    \begin{subfigure}{0.35\columnwidth}
    \centering
        \includegraphics[width=\textwidth]{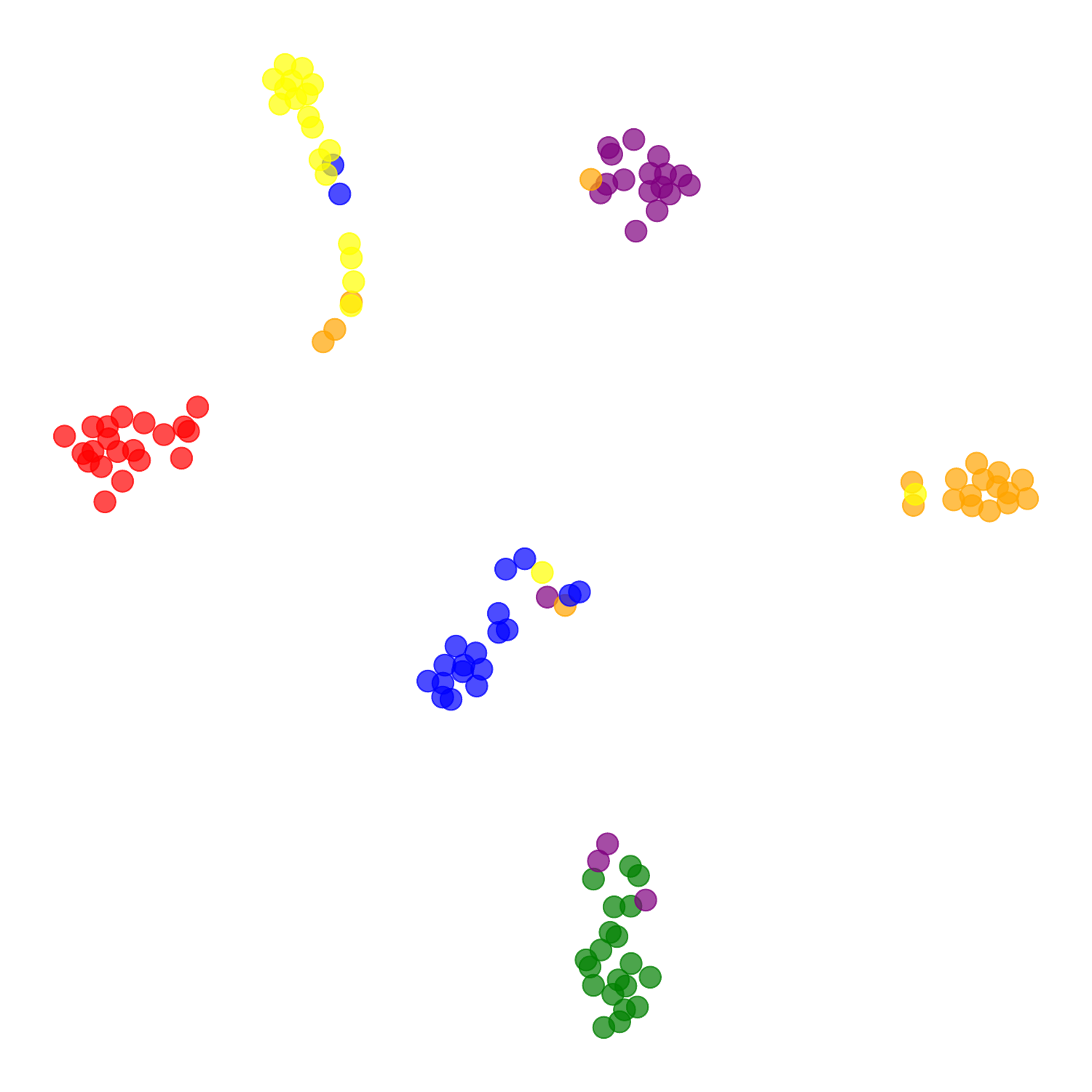}
        \caption{\tmodel}
        \label{fig:tsne_ood_vgg}
    \end{subfigure}
    \caption{t-SNE plots of the features inputted to the classifier head of the VGG16 backbone. Images with object occlusions of level L3 (60--80\%) from the Pascal test dataset are considered.}
    \label{fig:tsne_vgg}
\end{figure}

To analyse the improvements brought by \tmodel towards bridging the gap between
intermittent features of clean and occluded images of the same class, we
studied t-SNE plots of different types and levels of occlusions.
Fig.~\ref{fig:tsne_vgg} shows the t-SNE plots of the features (input to the
classifier head) of the Pascal test dataset with the highest level of
occlusion, L3 (60--80\%) for the VGG16 backbone. Due to high degree of
occlusions (missing/ noisy data), the intermittent features of the baseline
appear highly scattered, however with the \tmodel, these features are
well-clustered without decreasing the inter-class margins which in-turn
indicate the discriminative information of features. It should also be noted
that these features (output of \tmodel) are further subject to classification
layers of the backbone-- thus further enhancing the classification performance
of the proposed model as seen in Table~\ref{table:results_pascal_vgg}.

\noindent\textbf{Results on MS-COCO:} Table \ref{table:results_ms_coco_vgg}
shows the classification results on the MS-COCO dataset for the VGG16 backbone,
where the compared models were trained on the Pascal dataset. In addition to
the cross-dataset testing, this dataset provides another benefit of evaluating
the proposed method on realistic occlusion scenarios with varying degrees of
occlusions. Similar to the previous experiments, we ensure consistent protocols
to facilitate easy comparisons with the recent state-of-the-art methods of
classifying occluded images. Results of classification accuracy indicate that
the backbone incorporated with \tmodel outperforms all compared methods for
large occlusions (levels L2--L3) by at least 3.5\%. With an average accuracy of
92.1\%, our method provides over 13\% improvements in overall classification
over the baseline model. Similar to the experiments on Pascal dataset, the
proposed \tmodel enhances classification of clean data as well.

\begin{table}[h]
\renewcommand{\arraystretch}{1.3}
\centering
\begin{tabular}{|l|cccc|c|}
\hline
Occ. Area & \textbf{L0} & \textbf{L1} & \textbf{L2} & \textbf{L3} & \textbf{Mean} \\ \hline
baseline & 98.0 & 83.2 & 76.7 & 57.5 & 78.8 \\
CoD* & 91.8 & 82.7 & 83.3 & 76.7 & 83.6 \\
VGG+CoD* & 98.0 & 88.7 & 80.7 & 69.9 & 84.3 \\
TDAPNet* & 98.0 & 88.5 & 85.0 & 74.0 & 86.4 \\
CompNet* & 98.5 & \textbf{93.8} & 87.6 & 79.5 & 89.9 \\ \hline
Proposed & \textbf{99.2} & 93.3 & \textbf{91.1} & \textbf{84.9} & \textbf{92.1} \\ \hline
\end{tabular}
\caption{Performance evaluation of \tmodel on MS-COCO dataset using VGG16 backbone. All values refer to classification accuracy in \%. For the methods marked with *, we report the results from \cite{kortylewski2020compositional} for consistent protocol.}
\label{table:results_ms_coco_vgg}
\end{table}

Finally, we conduct the cross-dataset experiment (trained on Pascal, tested on
MS-COCO) with the ResNet-50 backbone. The comparative performance, as shown in
Table~\ref{table:results_ms_coco_resnet}, indicates that use of \tmodel brings
significant improvements over baseline for each level of occlusion as well as
on clean (unoccluded) images. With 89.9\% accurate classifications, the \tmodel
results in a boost of 9.3\% over the baseline, albeit falling short by 1\%
compared to the corresponding results obtained by the CompNet.

\begin{table}[h]
\renewcommand{\arraystretch}{1.3}
\centering
\begin{tabular}{|l|cccc|c|}
\hline
Occ. Area & \textbf{L0} & \textbf{L1} & \textbf{L2} & \textbf{L3} & \textbf{Mean} \\ \hline
baseline & 97.4 & 82.4 & 77.1 & 65.7 & 80.6 \\
CompNet~\cite{kortylewski2021compositional}& 98.5 & \textbf{92.6} & 88.9 & \textbf{83.6} & \textbf{90.9} \\ \hline
Proposed & \textbf{98.7} & 92.1 & \textbf{89.5} & 79.4 & 89.9 \\ \hline
\end{tabular}
\caption[]{Performance evaluation of \tmodel on MS-COCO dataset, using ResNet-50 backbone, on varying levels of occlusions. All values refer to classification accuracy in \%.}
\label{table:results_ms_coco_resnet}
\end{table}

%
\section{Conclusion}
\label{sec:conc}
In this work, we developed a method to enhance the classification of occluded images. Our model is an AE-based network that can be integrated into common CNN-based classification models in order to improve performance on occluded and OOD data while maintaining high accuracy on clean images. The novelty of our approach lies in devising loss functions across different layers of the overall architecture, which collectively contribute to the desired improvement. In addition to conventional losses used for AE, we introduce auxiliary intra-class loss to (partially) recover latent features of the occluded data. We also employ auxiliary inter-class loss in the same latent space to ensure compact representations of the class and thereby improve inter-class margin. Our experiments on two different datasets and two backbones show that the use of \tmodel brings as high as 25\% improvements over baseline for classification of highly occluded images without compromising classification of clean (in-distribution) images.

Although our current model successfully classifies a specific set of objects (12 in this case), it may become saturated and fail to scale for a larger number of classes. We are working towards combining multiple models in order to address this issue. 
%

\bibliographystyle{IEEEbib}
\bibliography{refs}

\end{document}